\documentclass[10pt,twocolumn,letterpaper]{article}
\usepackage[pagenumbers]{cvpr}
\usepackage{cvpr}  
\usepackage{makecell}
\usepackage{fp}
\usepackage{colortbl}
\usepackage{adjustbox}
\usepackage{amsfonts}
\usepackage{microtype}
\usepackage{float}
\usepackage[inline]{enumitem}
\usepackage[accsupp]{axessibility} 

\newcommand{\inlineheading}[1]{\vspace{1mm}\noindent\textbf{{#1}.}\hspace{0.5em}}

\newcommand{\mat}[1]{\mathrm{\textbf{#1}}}
\newcommand{\vect}[1]{\mathrm{\textbf{#1}}}

\newcommand{\fpmin}[5]{%
  \FPmin{\result}{#2}{#3}%
  \FPmin\result{\result}{#4}%
  \FPmin{#1}{\result}{#5}%
}
\newcommand{\fpmax}[5]{%
  \FPmax{\result}{#2}{#3}%
  \FPmax\result{\result}{#4}%
  \FPmax{#1}{\result}{#5}%
}

\def\fpnorm#1#2#3{%
\newcommand{#1}[1]{%
  \FPeval{\blendfactor}{(##1 - #2)/(#3 - #2)}%
  \FPeval{\xlt}{(\blendfactor * 2.0)}%
  \FPeval{\xgt}{(\blendfactor * 2.0) - 1.0}%
  \FPeval{\ylt}{(0.75)}%
  \FPeval{\ygt}{(1.0 - 0.75)}%
  \FPeval{\biaslt}{(\xlt / ((((1.0/\ylt) - 2.0)*(1.0 - \xlt))+1.0)) * 0.5}%
  \FPeval{\biasgt}{(\xgt / ((((1.0/\ygt) - 2.0)*(1.0 - \xgt))+1.0)) * 0.5 + 0.5}%
  \FPeval{\t}{\blendfactor}%
  \FPeval{\xt}{0.5}%
  \FPiflt\t\xt \FPset{\blendfactor}{\biaslt} \else \FPset{\blendfactor}{\biasgt} \fi%
  \FPeval{\blendred}{1-(\blendfactor-0.05)/(1.0-0.05)}%
  \FPmin{\blendred}{\blendred}{1.0}%
  \FPmax{\blendred}{\blendred}{0.0}%
  \FPset{\blendgreen}{\blendred}%
  \FPset{\blendblue}{1.0}%
  \FPset{\blendalpha}{0.25}%
  \FPeval{\blendblue}{ \blendalpha * \blendblue  + (1 - \blendalpha)}%
  \FPeval{\blendgreen}{\blendalpha * \blendgreen + (1 - \blendalpha)}%
  \FPeval{\blendred}{  \blendalpha * \blendred   + (1 - \blendalpha)}%
  
  \xdef\temp{\noexpand\cellcolor[rgb]{\blendred, \blendgreen, \blendblue}{##1}}
  \temp
  }%
}

\def\fpnorminv#1#2#3{%
\newcommand{#1}[1]{%
  \FPeval{\blendfactor}{1 - (##1 - #2)/(#3 - #2)}%
  \FPeval{\xlt}{(\blendfactor * 2.0)}%
  \FPeval{\xgt}{(\blendfactor * 2.0) - 1.0}%
  \FPeval{\ylt}{(0.85)}%
  \FPeval{\ygt}{(1.0 - 0.85)}%
  \FPeval{\biaslt}{(\xlt / ((((1.0/\ylt) - 2.0)*(1.0 - \xlt))+1.0)) * 0.5}%
  \FPeval{\biasgt}{(\xgt / ((((1.0/\ygt) - 2.0)*(1.0 - \xgt))+1.0)) * 0.5 + 0.5}%
  \FPeval{\t}{\blendfactor}%
  \FPeval{\xt}{0.5}%
  \FPiflt\t\xt \FPset{\blendfactor}{\biaslt} \else \FPset{\blendfactor}{\biasgt} \fi%
  \FPeval{\blendred}{1-(\blendfactor-0.5)/(1.0-0.5)}%
  \FPmin{\blendred}{\blendred}{1.0}%
  \FPmax{\blendred}{\blendred}{0.0}%
  \FPset{\blendgreen}{\blendred}%
  \FPset{\blendblue}{1.0}%
  \FPset{\blendalpha}{0.25}%
  \FPeval{\blendblue}{ \blendalpha * \blendblue  + (1 - \blendalpha)}%
  \FPeval{\blendgreen}{\blendalpha * \blendgreen + (1 - \blendalpha)}%
  \FPeval{\blendred}{  \blendalpha * \blendred   + (1 - \blendalpha)}%
  \xdef\temp{\noexpand\cellcolor[rgb]{\blendred, \blendgreen, \blendblue}{##1}}
  \temp
}%
}

\newcommand{\mynormcolor}[2]{%
\FPeval{\nc}{}%

}

\definecolor{cvprblue}{rgb}{0.21,0.49,0.74}
\usepackage[pagebackref,breaklinks,colorlinks,allcolors=cvprblue]{hyperref}

\def\paperID{14522} 
\def\confName{CVPR}
\def\confYear{2026}

\title{LaVR: Scene Latent Conditioned Generative Video Trajectory\\ Re-Rendering using Large 4D Reconstruction Models}

\author{ 
Mingyang Xie$^{1,2}$ \quad  Numair Khan$^{1}$    \quad Tianfu Wang$^{2}$  \quad Naina Dhingra$^{1}$  \quad Seonghyeon Nam$^{1}$ \quad Haitao Yang$^{1}$\\[0.5em]
Zhuo Hui$^{1}$ \quad Christopher Metzler$^{2}$ \quad Andrea Vedaldi$^{1,3}$ \quad Hamed Pirsiavash$^{1,4}$ \quad Lei Luo$^{1}$ \\ \\
$^{1}$Meta \quad $^{2}$ University of Maryland \quad $^{3}$University of Oxford \quad $^{4}$UC Davis\\\\
\href{https://lavr-4d-scene-rerender.github.io/}{\textcolor{magenta}{lavr-4d-scene-rerender.github.io}} 
}

\begin{document}

\twocolumn[{
\maketitle

\vspace{-17pt}  

\centering
\includegraphics[width=\textwidth]{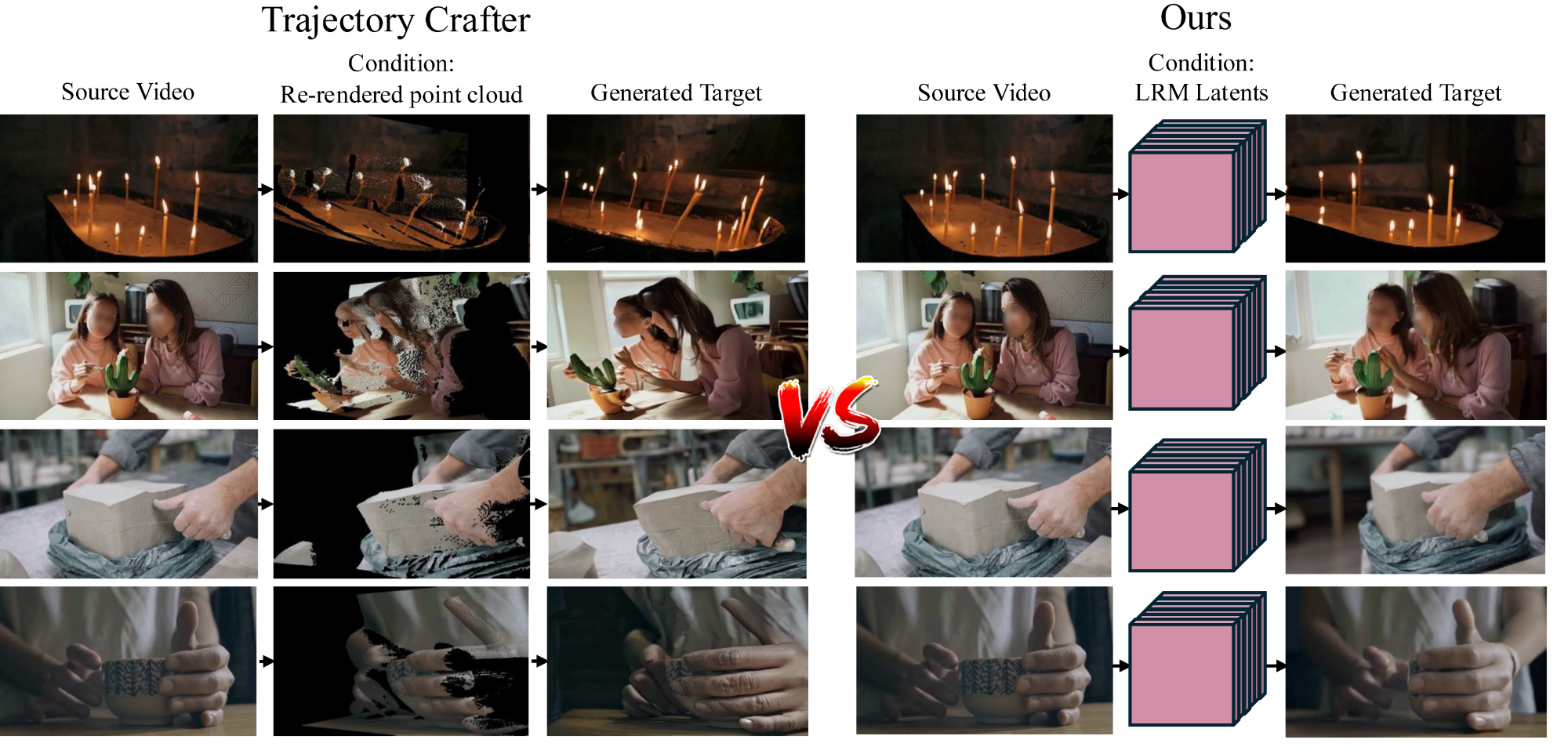}

\captionof{figure}{
Our method addresses the problem of rendering geometrically consistent novel trajectories from a monocular source video.
We propose to utilize the geometric knowledge of a pretrained large reconstruction model (LRM) by conditioning the trajectory generation process on the latent state of a 4D LRM.
Compared to prior methods that are conditioned on error-prone point cloud re-renderings of the source video, our method achieves state-of-the-art visual quality while maintaining a high level of geometric fidelity to the original scene.
}

\label{fig:teaser}
\par\vspace{17pt}  
}]






\begin{abstract}
Given a monocular video, the goal of video re-rendering is to generate views of the scene from a novel camera trajectory.
Existing methods face two distinct challenges.
Geometrically unconditioned models lack spatial awareness, leading to drift and deformation under viewpoint changes.
On the other hand, geometrically conditioned models depend on estimated depth and explicit reconstruction, making them susceptible to depth inaccuracies and calibration errors.
\\\\
We propose to address these challenges by using the implicit geometric knowledge embedded in the latent space of a large 4D reconstruction model to condition the video generation process.
These latents capture scene structure in a continuous space without explicit reconstruction.
Therefore, they provide a flexible representation that allows the pretrained diffusion prior to regularize errors more effectively.
By jointly conditioning on these latents and camera poses, we demonstrate that our model achieves state-of-the-art results on the video re-rendering task.
\end{abstract}    
\section{Introduction}%
\label{sec:intro}

Video re-rendering, or dynamic novel view synthesis, aims to visualize a dynamic scene from new and unseen camera paths.
Unlike standard video generation, this task requires modeling both scene dynamics and underlying geometry to maintain temporal and spatial coherence under arbitrary camera motion.
Video re-rendering enables various applications, including re-rendering captured scenes for film and generating immersive experiences from a single video.
However, the problem is inherently challenging.
Monocular inputs provide weak geometric supervision, forcing models to jointly infer structure, motion, and lighting.

Existing works approach this problem from two directions.
Geometrically conditioned methods~\cite{mark2025trajectorycrafter,ren2025gen3c,gu2025diffusion_as_shader} model scene geometry using point clouds or meshes and re-render novel views.
Although physically grounded, these methods~\cite{mark2025trajectorycrafter, ren2025gen3c} rely on accurate depth estimation, so any errors propagate into the re-rendered point clouds and cause shape distortions in novel views, where objects appear stretched or compressed along the depth direction and parallax becomes inconsistent.
To address this issue, geometrically unconditioned methods~\cite{bai2025recammaster} have been recently proposed to avoid explicit conditioning and instead generate videos using only the input video and target trajectory.
Such methods achieve strong visual realism---largely inherited from the pretrained video diffusion prior---but struggle with geometric consistency across viewpoints.
This motivates a new approach that can capture the strengths of both directions by providing geometric guidance without depending on precise depth.

We propose a generative video trajectory re-rendering model that conditions a video diffusion backbone on a latent 4D scene representation extracted from monocular videos.
Instead of explicit 4D conditioning, we encode input videos into a latent space that captures appearance, geometry, and dynamics, enabling the model to follow novel camera trajectories while maintaining coherent structure and parallax.

This formulation is enabled by recent large 4D reconstruction models~\cite{cut3r,wang2025vggt,li2025megasam} (LRMs), which have shown that a feed-forward network can extract rich latent representations from monocular frames and decode them into depth, pose, or approximate novel views.
These models demonstrate that implicit geometry structure can be captured without explicit optimization or volumetric reconstruction, providing exactly the type of geometry-aware cues our framework leverages.

While both point clouds and LRM-produced latents can provide geometric cues, they condition the generative process in fundamentally different ways.
Point-cloud pipelines reconstruct geometry from estimated depth and re-render it from the target viewpoint, so any depth error directly manifests as distorted shapes, incorrect parallax, or missing regions---acting as a rigid geometric constraint that leaves the generative model little flexibility to correct mistakes.
In contrast, implicit geometry latents provide structural guidance in a softer, non-pixel-aligned form.
As diffusion models are pretrained on large-scale video data with strong priors over plausible motion and scene structure, they can regularize small geometric inconsistencies in these latents.
This combination yields geometric cues that are both informative and robust to depth noise, motivating our design choice.

\begin{figure}
\centering
\includegraphics[trim=0cm 0cm 0cm 0cm, clip=true, width=1.0\linewidth]{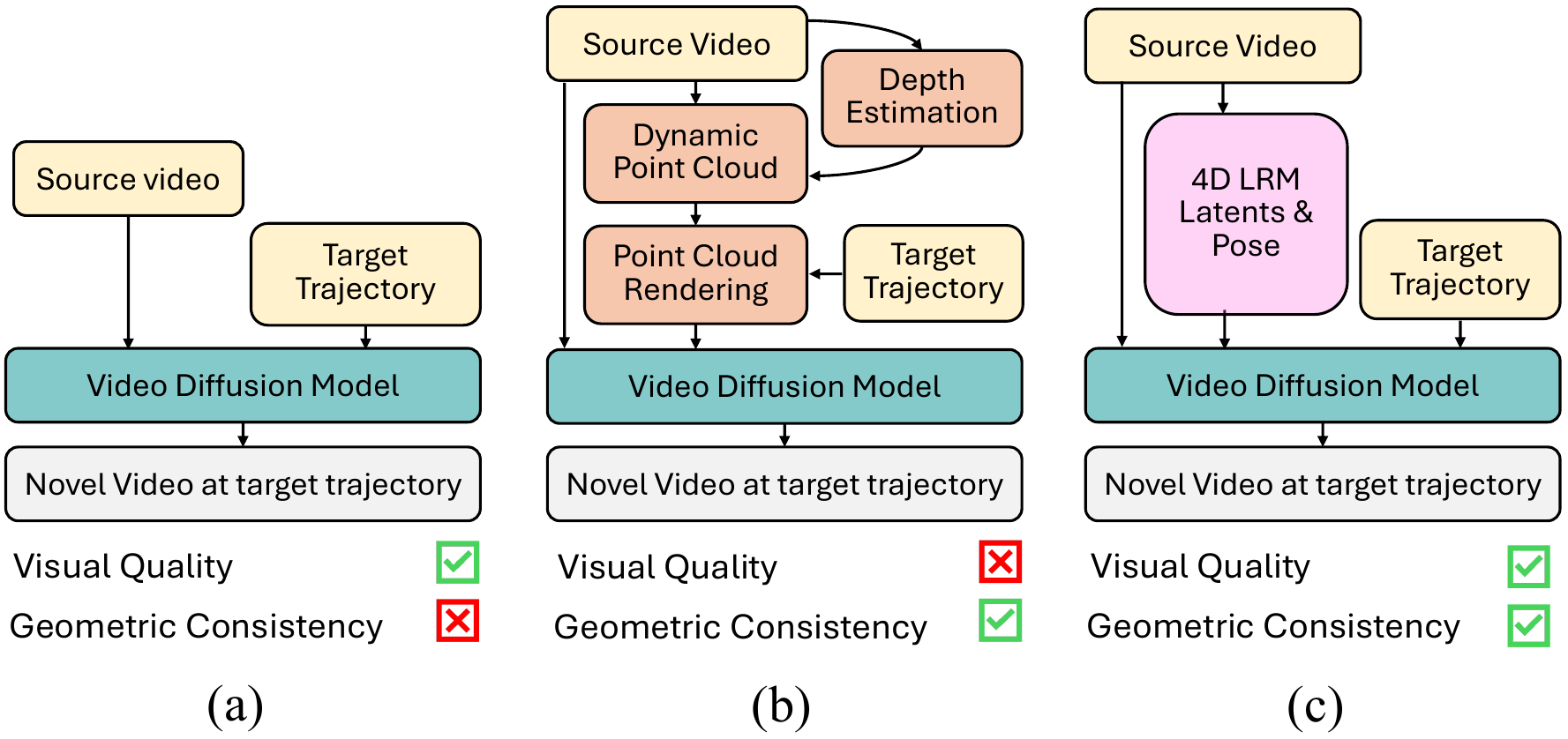}
\caption{\textbf{Architecture comparison.}
\textbf{(a)} Unconditioned methods for novel trajectory generation achieve high visual quality but lack geometric awareness, leading to inconsistencies.
\textbf{(b)} Conditioning on 4D point cloud renders provides consistency but reduces quality because the depth estimation and point cloud generation stages are sensitive to errors.
\textbf{(c)} Our proposed architecture utilizes the implicit geometric knowledge of a pre-trained large 4D reconstruction model (LRM) to achieve both high quality and consistency.}%
\label{fig:paradigm_comparison}
\end{figure}

As a result, our latent space conditioning formulation provides strong geometric priors without relying on accurate depth estimation.
It enables the model to maintain stable parallax and coherent structure under large viewpoint changes, producing geometrically consistent videos along arbitrary camera trajectories.
\Cref{fig:teaser} illustrates the qualitative advantages of our approach compared with a state-of-the-art explicitly 4D-conditioned method~\cite{mark2025trajectorycrafter}, and \cref{fig:paradigm_comparison} compares our overall paradigm with existing baselines.
In summary, our novel contributions are:
\begin{itemize}
\item We propose to use the latent state of a large reconstruction model to provide geometric conditioning without relying on explicit depth or point cloud reconstruction.

\item We present a lightweight adapter module that compresses and integrates the latents from a state-of-the-art 4D reconstruction model with VAE-encoded video latents for efficient consumption by a pretrained diffusion backbone.

\item We conduct extensive evaluation to show that our approach outperforms both geometrically-conditioned and unconditioned baselines on quantitative and qualitative metrics.
\end{itemize}

\begin{figure*}[t]
\centering

\includegraphics[width=1.0\linewidth,trim=0cm 0.0cm 0cm 0.0cm,clip]{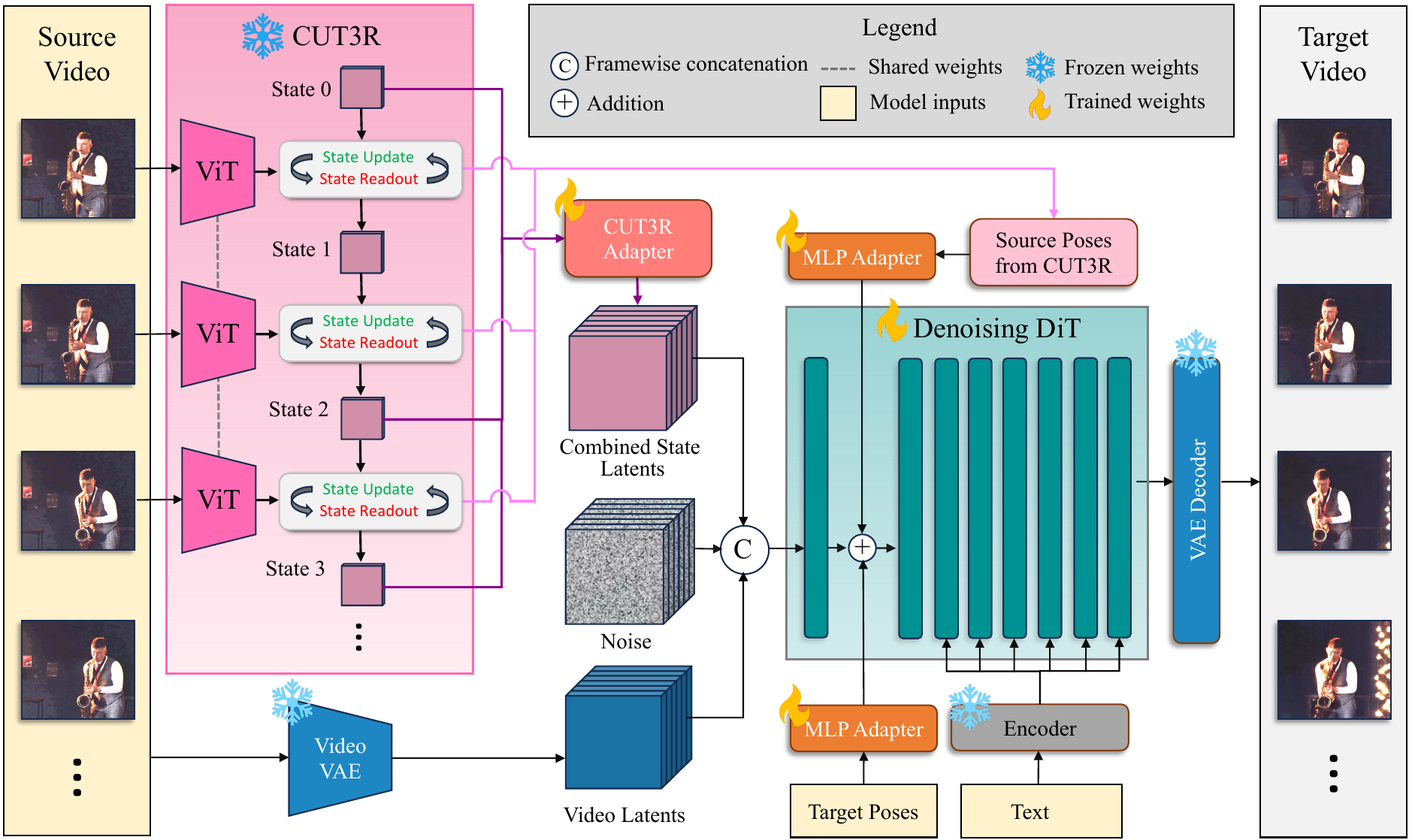}

\caption{\textbf{Pipeline overview.}
Given a monocular source video, our method generates a novel video of the same scene at a target camera trajectory using a video diffusion model.
To ensure geometric consistency, we condition the model on latents from CUT3R~\cite{cut3r}, a pretrained 4D reconstruction model.
We use four signals from the source video: the standard video VAE latents, CUT3R's 4D latents, source camera poses, and an encoded text description of the scene.
A novel adapter architecture aligns the CUT3R and VAE latents and allows these to be fed to the model in a computationally feasible manner.
The source camera poses come from CUT3R and are added to the DiT's intermediate activations after passing through a small MLP-based adapter.
Another MLP processes the target poses at which the novel video is rendered.
Note that only the projection and self-attention layers of the DiT are trainable, while other parameters are frozen.}%
\label{fig:pipeline}
\end{figure*}

\section{Related Work}%
\label{sec:related-work}

\inlineheading{Unsupervised 3D/4D Scene Reconstruction}
Neural scene representations such as NeRF~\cite{Mildenhall2020NeRF, barron2022mip, barron2021mip} and 3D Gaussian Splatting~\cite{Kerbl20233DGS, huang20242d, yu2024mip, zhu2023FSGS, xie24flashsplat} have advanced novel view synthesis by reconstructing scene geometry from posed images.
While effective for static scenes with sufficient multi-view coverage, their reliance on accurate and complete geometry makes them difficult to apply when only monocular video frames are available.
Subsequent work has extended these approaches to dynamic scenes~\cite{nerfies, song2023nerfplayer, dynamic_nerf, pumarola2021d, Wu_2024_CVPR_4dgs, duan20244d, som2024shape_from_motion}.
Early monocular methods rely on depth-based warping~\cite{yoon2020novel, ke2025marigold} with later improvements in occlusion handling, while newer techniques adopt neural dynamic representations~\cite{nerfies, song2023nerfplayer, dynamic_nerf, pumarola2021d} or Gaussian Splatting variants~\cite{Wu_2024_CVPR_4dgs, duan20244d,som2024shape_from_motion} with additional regularization to stabilize time-varying geometry.
Nonetheless, these pipelines still often degrade when the target camera trajectory departs significantly from the observed views.

\inlineheading{Large 3D/4D Reconstruction Models}
Large 3D and 4D reconstruction models~\cite{wang2023dust3r, wang2024spann3r, cut3r, wang2025vggt, monst3r} leverage high-capacity architectures and large-scale pretraining to estimate scene structure from single images, image pairs, or short video clips.
Early feed-forward approaches such as DUST3R~\cite{wang2023dust3r} demonstrate that transformer-based correspondence aggregation alone can recover camera poses and dense geometry without iterative optimization.
Building on this idea, more recent systems such as SPANN3R~\cite{wang2024spann3r}, CUT3R~\cite{cut3r}, MONST3R~\cite{monst3r}, and MegaSAM~\cite{li2025megasam} scale model capacity and training data to achieve more generalizable reconstructions and capture time-varying or 4D scene structure.
These large reconstruction models provide strong geometric priors that can serve as effective conditioning signals for generative and camera-controlled video synthesis.

\inlineheading{Generative Novel View Synthesis}
Recent works in this direction have explored using generative models to achieve controllable camera trajectories for video synthesis~\cite{you2024nvs, yu2024viewcrafter, muller2024multidiff, liu2024reconx, sun2024dimensionx, ren2025gen3c, hu2025ex4dextremeviewpoint4d, wang2023breathing, bahmani2025ac3d, bahmani2024vd3d, wang2025flash, bai2024syncammaster, he2024cameractrl, hou2024training, van2024generative, xu2024camco, wang2024motionctrl, ren2024customize}.
These methods typically condition video diffusion models on camera poses or trajectory signals to guide viewpoint changes.
In the dynamic setting, TrajectoryCrafter~\cite{mark2025trajectorycrafter}, Gen3C~\cite{ren2025gen3c}, and EX-4D~\cite{hu2025ex4dextremeviewpoint4d} generate new views by conditioning on rendered point clouds or meshes, offering strong consistency but inheriting depth-related brittleness.
By contrast, ReCamMaster~\cite{bai2025recammaster} directly synthesizes videos along new trajectories without explicit 4D structure, providing greater flexibility at the cost of weaker geometric stability.

Explicit 4D pipelines therefore excel at enforcing geometry but are vulnerable to reconstruction errors, whereas non-4D-conditioned methods remain more robust but struggle to maintain spatial coherence under large camera motion.

\section{Method}%
\label{sec:method}

Given a source video, our goal is to synthesize novel frames along a user-specified camera trajectory while preserving the scene content and dynamics of the source.
To achieve this, we propose conditioning a video-to-video diffusion model on the latent state of a large 4D reconstruction model (LRM).

Our proposed approach exploits the fact that the latent state aggregates scene structure and camera motion without explicit geometric reconstruction.
Thus, it circumvents the error-prone approach of depth estimation and point-cloud/mesh reconstruction adopted by existing models with geometric conditioning~\cite{mark2025trajectorycrafter, ren2025gen3c, wu2025spmem}.
In addition, these prior works utilize geometry only indirectly through renderings.
These renderings---which suffer from distortions and holes---bake the reconstructed scene into a 2D image and leave the generative model limited room to reason about the underlying geometry and correct errors.
In contrast, the latent space of an LRM encodes geometry and camera poses in a high-dimensional feature space.
Not only does this preserve the entire 4D scene structure, the continuous representation allows more flexibility for the pre-trained video diffusion prior to regularize inconsistencies during the generation stage.

In addition to the latent state, we use the source camera poses and a text prompt as secondary conditions to provide additional context for the input frames.
The user-specified target poses are supplied as a control signal to steer the denoising process toward the desired camera path.
The source and target poses are injected into each DiT block via two separate lightweight MLP adapters, enabling effective conditioning on camera trajectories.

\begin{figure}
\centering
\includegraphics[ width=1.0\linewidth]
{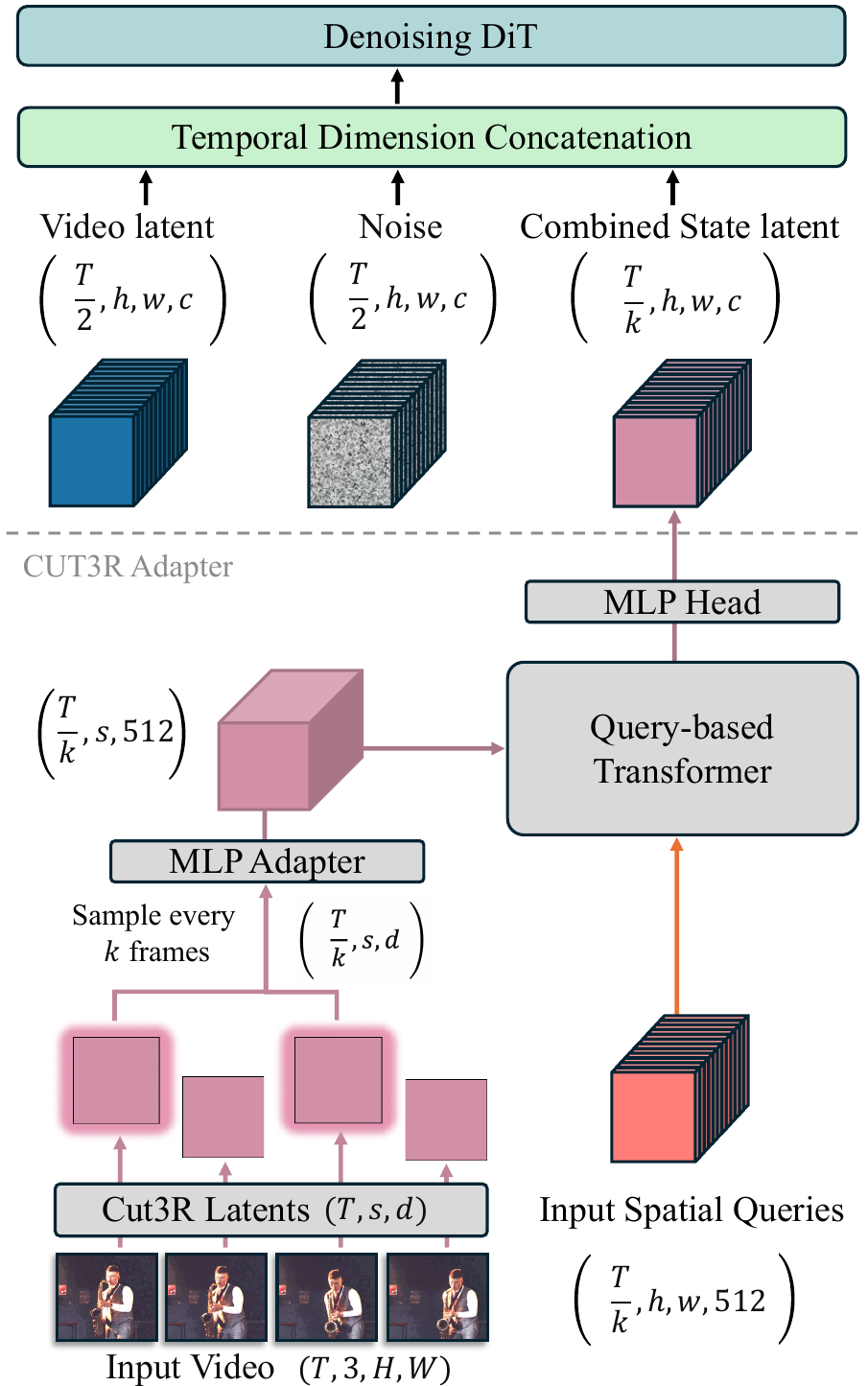}
\caption{\textbf{Proposed CUT3R Adapter.}
Our lightweight adapter embeds CUT3R's per-frame latent tokens into geometry-aware features that align with the representation used by the diffusion model.
The shape of features at each stage is shown in brackets.}%
\label{fig:cut3r-adapter}
\end{figure}
\subsection{Scene State Latents with CUT3R}

We use CUT3R~\cite{cut3r} as a representative 4D LRM for our task.
CUT3R provides consistent feed-forward reconstruction from a monocular video by maintaining a persistent latent state that aggregates multi-view information over time and reflects the evolving 3D understanding of the scene.
Multiple heads decode the state to recover the pose, world-space point maps, and depth for each frame of the input video, thereby demonstrating that the latent state contains strong geometry and motion cues for downstream conditioning.

The latent state is represented as a set of $s$ tokens $\{\ell^i \in \mathbb{R}^d\}_{i=1}^s$ that are updated at each time step by the ViT-encoded frame of the source video (see \cref{fig:pipeline}).
We use the state tokens $\mathcal{S} = \{ \{ \ell^i_t \in \mathbb{R}^d\}_{i=1}^s, t=1, 2, \dots, T\}$ for all $T$ frames of the video to preserve temporal changes in both scene content and camera pose.

\subsection{Adapting CUT3R Latents to the DiT Backbone}

While the CUT3R state latents $S$ provide rich geometric cues, they are represented as token-based implicit scene states rather than spatial video latents, and are therefore not directly compatible with the input interface of the diffusion backbone. We therefore introduce a lightweight adapter that transforms frame-wise CUT3R tokens into geometry-aware spatial features that can be consumed by the pretrained denoising diffusion transformer (DiT). Its overall framework is illustrated in \cref{fig:cut3r-adapter}.

Starting from the CUT3R state latent tensor $S$ of shape $(T, s, d)$, we first subsample the states every $k$-th frame, yielding $T/k$ token sets. The sampled tokens are embedded by an MLP adapter and then processed by a query-based transformer that converts each token set into a spatial feature map. Specifically, we use a set of spatial queries corresponding to the target $h \times w$ grid and apply cross-attention from these queries to CUT3R tokens. This produces a spatial feature map for each sampled CUT3R frame, where each output location aggregates information from the full token set.

After the query-based transformer, the resulting spatial features are further projected to match the channel dimensionality of the video VAE latents, producing geometry-aware latent features of shape $(T/k, h, w, c)$, where $c$ is the channel dimension of the video VAE latents. We then concatenate these adapted CUT3R features with the source video latents and noisy output video latents along the temporal dimension before feeding them into the DiT. This allows us to inject geometry-aware conditioning without modifying the backbone architecture, while remaining compatible with the pretrained spatiotemporal organization of the diffusion model. 

\subsection{Training Strategy}%
\label{sec:3d}

We train the CUT3R adapter and the two MLP pose adapters from scratch and finetune a subset of DiT layers.
These layers include the DiT projector and all self-attention blocks.
The remaining layers of the DiT and the Video VAE are frozen to preserve pretrained priors.
We train the model on the synthetic MultiCamVideo dataset from ReCamMaster~\cite{bai2025recammaster}, which provides multiple posed trajectories for dynamic scenes.
We randomly select two trajectories per scene as source and target.

We use a standard conditional flow-matching loss~\cite{Lipman2022FlowMF} as the training objective.
Specifically, given the clean \emph{target} latent $z_0$, a noise sample 
$\epsilon \sim \mathcal{N}(0,I)$,
and a diffusion timestep
$t \sim \mathcal{U}(0,1)$, 
we set the interpolated latent $z_t = (1-t)z_0 + t\epsilon$. 
The DiT predicts the velocity field conditioned on the adapted CUT3R state latents $Z_{c}$ 
and the source video latents $Z_{s}$ using the following objective:
\[
\mathcal{L}_{\text{FM}}
=
\mathbb{E}_{t,\,z_0,\,\epsilon}
\left[
\left \|
v_\theta(z_t,\, t,\, Z_{c},\, Z_{s})
\;-\;
(\epsilon - z_0)
\right \|_2^2
\right].
\]
To allow fast convergence of the geometry-conditioned pathway and preserve pretrained priors, we use a 3$\times$ higher learning rate for the CUT3R adapter than other components.

\begin{figure}
\centering
\includegraphics[width=1.0\linewidth]{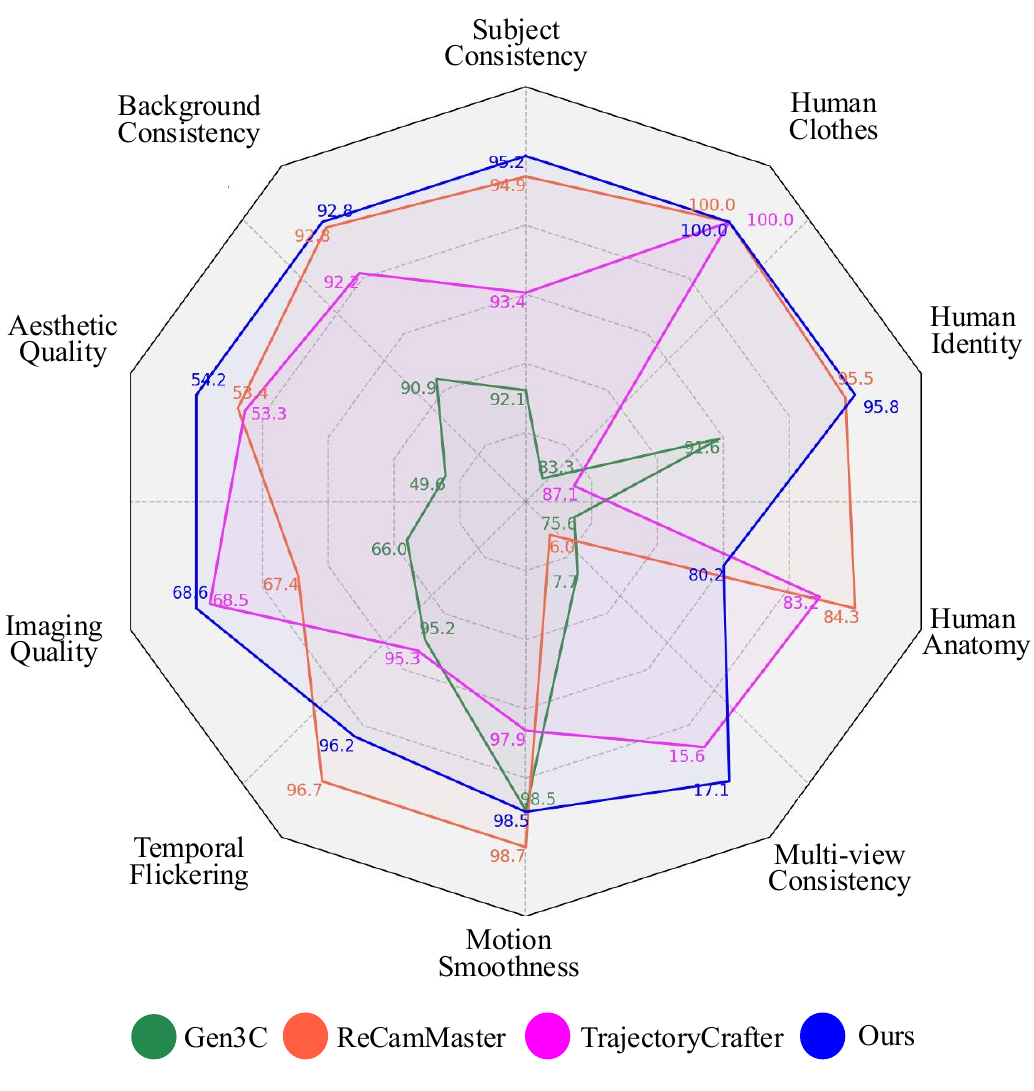}
\caption{\textbf{Evaluation on the VBench~\cite{huang2023vbench, zheng2025vbench2} metrics.}
We highlight relative differences by normalizing each metric over all baselines.
Our method shows all-around high performance, achieving the best results for multi-view, subject, and background consistency.}%
\label{fig:vbench-metrics}
\end{figure}

\begin{figure*}[t]
\centering
\includegraphics[width=1.0\linewidth]{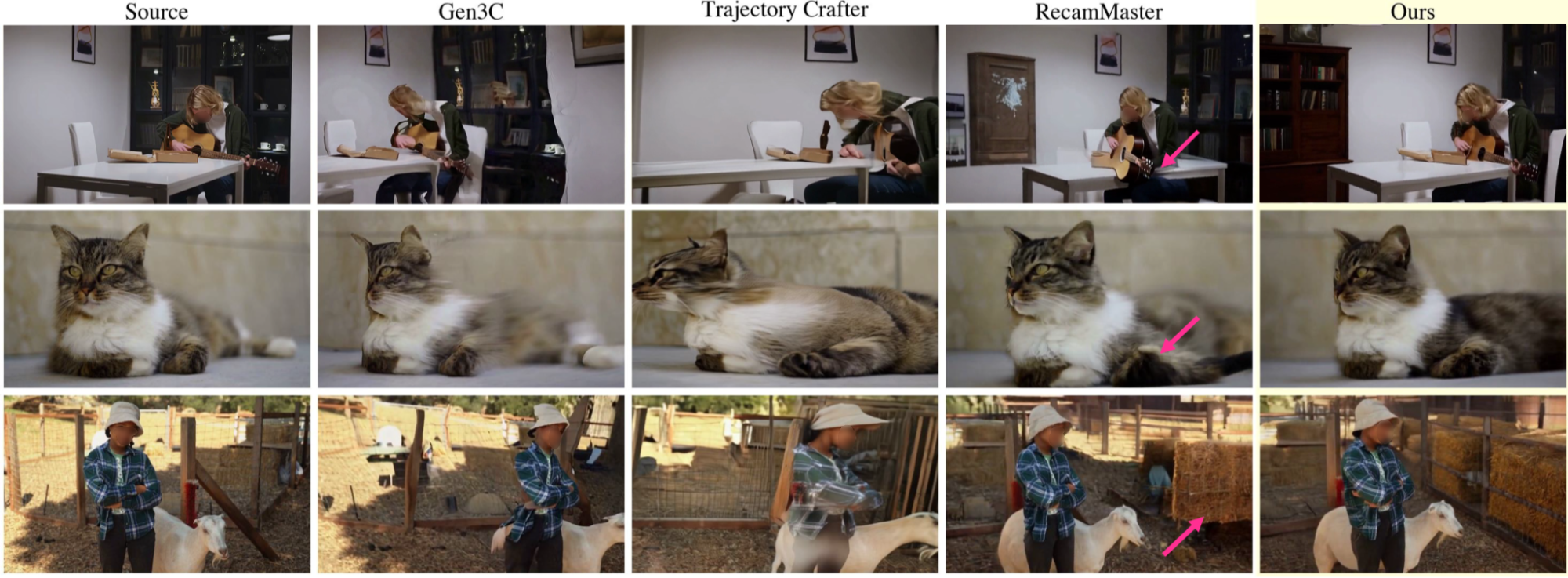}
\caption{\textbf{Qualitative evaluation of novel views.}
We compare frames from new camera trajectories rendered by redirecting the source video.
Both Gen3C~\cite{ren2025gen3c} and TrajectoryCrafter~\cite{mark2025trajectorycrafter} are conditioned on re-rendered point clouds and suffer from unnatural warping artifacts.
ReCamMaster~\cite{bai2025recammaster} is not geometrically conditioned and hallucinates implausible content in unseen regions (\textit{top:} man overlaps with table; \textit{middle:} cat's tail; \textit{bottom:} haystacks in the background).
Compared to baselines, our results look natural and geometrically consistent.}%
\label{fig:qualitative-all-methods}
\end{figure*}
\def\psnrgenthreec{20.62}
\def\psnrtrajcrafter{14.84}
\def\psnrrecam{17.75}
\def\psnrours{20.74}

\def\ssimgenthreec{}
\def\ssimtrajcrafter{}
\def\ssimrecam{}
\def\ssimours{}

\def\lpipsgenthreec{23.23}
\def\lpipstrajcrafter{41.59}
\def\lpipsrecam{32.63}
\def\lpipsours{22.47}

\def\clipgenthreec{97.47}
\def\cliptrajcrafter{95.05}
\def\cliprecam{97.03}
\def\clipours{98.07}

\def\subjectgenthreec{92.07}
\def\subjecttrajcrafter{93.38}
\def\subjectrecam{94.95}
\def\subjectours{95.22}

\def\multiviewgenthreec{7.695}
\def\multiviewtrajcrafter{15.57}
\def\multiviewrecam{5.975}
\def\multiviewours{17.11}

\def\backgroundgenthreec{90.91}
\def\backgroundtrajcrafter{92.21}
\def\backgroundrecam{92.76}
\def\backgroundours{92.83}

\fpmin{\psnrmin}{\psnrgenthreec}{\psnrtrajcrafter}{\psnrrecam}{\psnrours}
\fpmax{\psnrmax}{\psnrgenthreec}{\psnrtrajcrafter}{\psnrrecam}{\psnrours}
\fpnorm{\psnr}{\psnrmin}{\psnrmax}

\fpmin{\ssimmin}{\ssimgenthreec}{\ssimtrajcrafter}{\ssimrecam}{\ssimours}
\fpmax{\ssimmax}{\ssimgenthreec}{\ssimtrajcrafter}{\ssimrecam}{\ssimours}
\fpnorm{\ssim}{\ssimmin}{\ssimmax}

\fpmin{\lpipsmin}{\lpipsgenthreec}{\lpipstrajcrafter}{\lpipsrecam}{\lpipsours}
\fpmax{\lpipsmax}{\lpipsgenthreec}{\lpipstrajcrafter}{\lpipsrecam}{\lpipsours}
\fpnorminv{\lpips}{\lpipsmin}{\lpipsmax}

\fpmin{\clipmin}{\clipgenthreec}{\cliptrajcrafter}{\cliprecam}{\clipours}
\fpmax{\clipmax}{\clipgenthreec}{\cliptrajcrafter}{\cliprecam}{\clipours}
\fpnorm{\clip}{\clipmin}{\clipmax}

\fpmin{\subjectmin}{\subjectgenthreec}{\subjecttrajcrafter}{\subjectrecam}{\subjectours}
\fpmax{\subjectmax}{\subjectgenthreec}{\subjecttrajcrafter}{\subjectrecam}{\subjectours}
\fpnorm{\subject}{\subjectmin}{\subjectmax}

\fpmin{\multiviewmin}{\multiviewgenthreec}{\multiviewtrajcrafter}{\multiviewrecam}{\multiviewours}
\fpmax{\multiviewmax}{\multiviewgenthreec}{\multiviewtrajcrafter}{\multiviewrecam}{\multiviewours}
\fpnorm{\multiview}{\multiviewmin}{\multiviewmax}

\fpmin{\backgroundmin}{\backgroundgenthreec}{\backgroundtrajcrafter}{\backgroundrecam}{\backgroundours}
\fpmax{\backgroundmax}{\backgroundgenthreec}{\backgroundtrajcrafter}{\backgroundrecam}{\backgroundours}
\fpnorm{\background}{\backgroundmin}{\backgroundmax}

\setlength{\tabcolsep}{5.5pt}
\begin{table*}[t!]
\begin{center}
\begin{tabular}{lccccccc}
\toprule
& \multicolumn{3}{c}{Cycle Consistency} & \multicolumn{4}{c}{VBench Consistency} \\
\midrule
Method & PSNR$\uparrow$ & LPIPS$\downarrow$ & CLIP$\uparrow$ & & Subject$\uparrow$ & Multi-view$\uparrow$ & Background$\uparrow$\\
\midrule
Gen3C~\cite{ren2025gen3c}& \psnr{\psnrgenthreec} & \lpips{\lpipsgenthreec} & \clip{\clipgenthreec} & & \subject{\subjectgenthreec} & \multiview{\multiviewgenthreec} & \background{\backgroundgenthreec}\\
TrajectoryCrafter~\cite{mark2025trajectorycrafter} & \psnr{\psnrtrajcrafter} & \lpips{\lpipstrajcrafter} & \clip{\cliptrajcrafter} & & \subject{\subjecttrajcrafter} & \multiview{\multiviewtrajcrafter}& \background{\backgroundtrajcrafter}\\
ReCamMaster~\cite{bai2025recammaster} & \psnr{\psnrrecam} & \lpips{\lpipsrecam} & \clip{\cliprecam} & & \subject{\subjectrecam} & \multiview{\multiviewrecam} & \background{\backgroundrecam}\\
Ours & \textbf{\psnr{\psnrours}} & \textbf{\lpips{\lpipsours}} & \textbf{\clip{\clipours}} & & \textbf{\subject{\subjectours}} & \textbf{\multiview{\multiviewours}} & \textbf{\background{\backgroundours}} \\
\midrule
\bottomrule
\end{tabular}
\end{center}

\vspace{-10pt}

\caption{\textbf{Quantitative Comparison of Consistency.}
We highlight the metrics in blue, proportional to their percentile.
The values for LPIPS, CLIP, and all VBench metrics are $\times 10^{-2}$.
Our method shows strong performance on all metrics, achieving the best results on cycle consistency.
Please refer to the qualitative results in \cref{fig:qualitative-recam-errors,fig:qualitative-all-grid} and to the supplementary material for video results.}%
\label{table:quantitative-results-c}
\end{table*}
\def\abstgenthreec{24.45}
\def\absttrajcrafter{16.53}
\def\abstrecam{21.83}
\def\abstours{14.39}

\def\reltgenthreec{12.00}
\def\relttrajcrafter{10.52}
\def\reltrecam{12.43}
\def\reltours{7.798}

\def\relrgenthreec{0.641}
\def\relrtrajcrafter{0.442}
\def\relrrecam{0.518}
\def\relrours{0.411}

\fpmin{\abstmin}{\abstgenthreec}{\absttrajcrafter}{\abstrecam}{\abstours}
\fpmax{\abstmax}{\abstgenthreec}{\absttrajcrafter}{\abstrecam}{\abstours}
\fpnorminv{\abst}{\abstmin}{\abstmax}

\fpmin{\reltmin}{\reltgenthreec}{\relttrajcrafter}{\reltrecam}{\reltours}
\fpmax{\reltmax}{\reltgenthreec}{\relttrajcrafter}{\reltrecam}{\reltours}
\fpnorminv{\relt}{\reltmin}{\reltmax}

\fpmin{\relrmin}{\relrgenthreec}{\relrtrajcrafter}{\relrrecam}{\relrours}
\fpmax{\relrmax}{\relrgenthreec}{\relrtrajcrafter}{\relrrecam}{\relrours}
\fpnorminv{\relr}{\relrmin}{\relrmax}

\setlength{\tabcolsep}{5.5pt}
\begin{table}[t!]
\begin{center}
\begin{tabular}{lccc}
\toprule
& \multicolumn{3}{c}{Pose Reconstruction Error} \\
\midrule
Method & $\mathrm{Abs}(\vect{t})\downarrow$ &  $\mathrm{Rel}(\vect{t})\downarrow$ & $\mathrm{Rel}(\mat{R})\downarrow$ \\
\midrule
Gen3C & \abst{\abstgenthreec} & \relt{\reltgenthreec} & \relr{\relrgenthreec} \\
TrajectoryCrafter & \abst{\absttrajcrafter} & \relt{\relttrajcrafter} & \relr{\relrtrajcrafter} \\
ReCamMaster & \abst{\abstrecam} & \relt{\reltrecam} & \relr{\relrrecam} \\
            Ours & \textbf{\abst{\abstours}} & \textbf{\relt{\reltours}} & \textbf{\relr{\relrours}} \\
\midrule
\bottomrule
\end{tabular}
\end{center}
\vspace{-10pt}
\caption{\textbf{Target pose reconstruction accuracy.}
We evaluate the absolute $\mathrm{Abs}(\cdot)$ and relative $\mathrm{Rel}(\cdot)$ errors in camera translation $\vect{t}$ (in millimeters) and rotation $\mat{R}$ (in degrees).
While we achieve consistently high rotational and translational accuracy, the unconditioned ReCamMaster~\cite{bai2025recammaster} fails to follow the target trajectory closely.
We highlight the metrics in blue, proportional to their percentile.}%
\label{table:quantitative-results-d}
\end{table}
\begin{figure*}[t]
\centering
\includegraphics[width=1.0\linewidth]{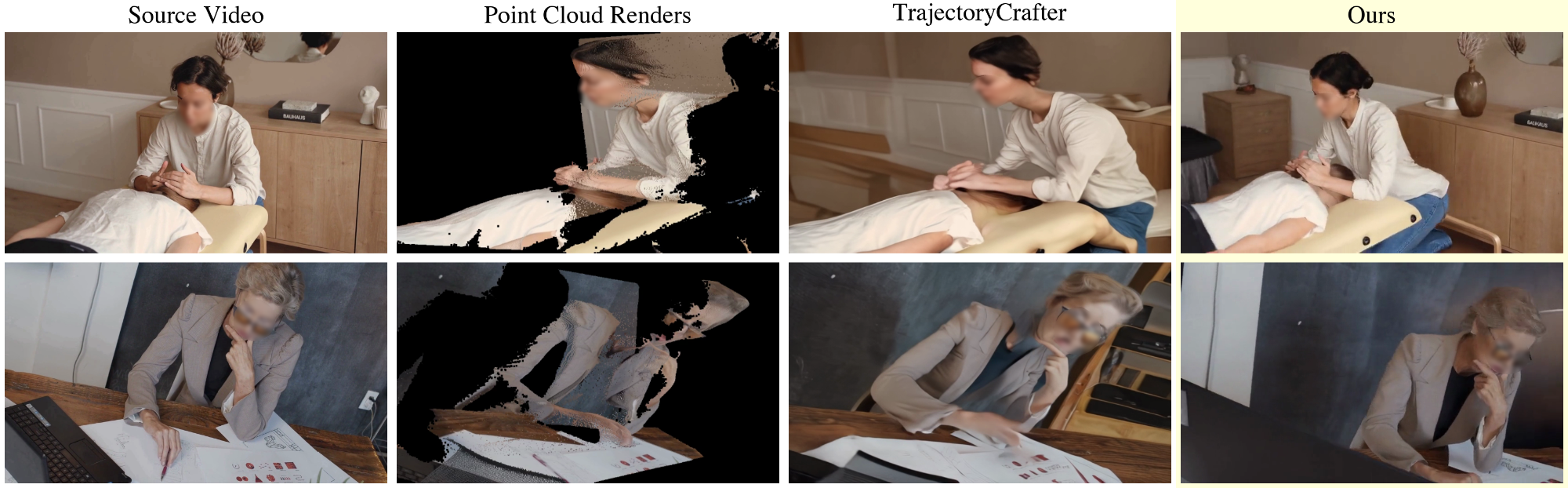}
\caption{\textbf{Disadvantages of geometric conditioning via re-rendered point clouds.}
Visualizing the point cloud renders of TrajectoryCrafter, we see that depth scale ambiguity, empirically estimated intrinsics, and holes or misalignment errors in the point cloud can create warped conditioning images that lead to unnatural outputs.
Our results do not suffer from such artifacts.}%
\label{fig:qualitative-trajcrafter-errors}
\end{figure*}
\begin{figure*}[t]
\centering
\includegraphics[width=1.0\linewidth]{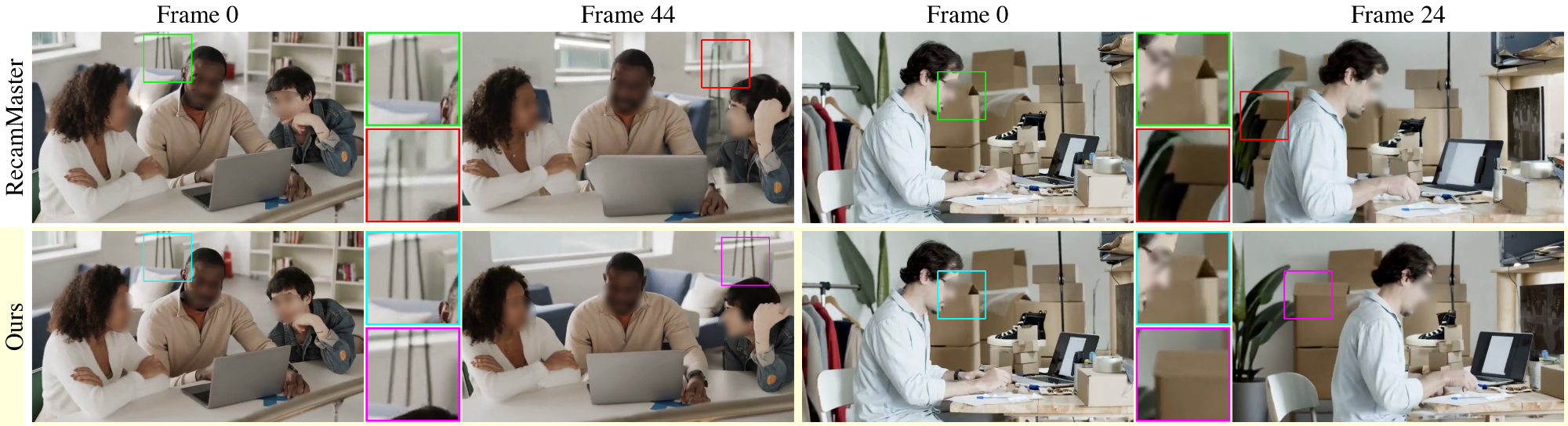}
\caption{\textbf{Qualitative evaluation of geometric consistency across frames.}
We show the beginning and end frames for two novel video trajectories from ReCamMaster~\cite{bai2025recammaster} and our method.
Lacking any geometric conditioning, ReCamMaster fails to maintain consistency across occlusions (\textit{Left:} a leg of the lamp vanishes; \textit{Right:} the cardboard box opens after reappearing behind the subject).
The latent 4D condition of our method produces a more stable reconstruction.
Please refer to the supplementary material for video results.}%
\label{fig:qualitative-recam-errors}
\end{figure*}
\setlength{\tabcolsep}{3.5pt}

\begin{table}[h]
\begin{center}
\resizebox{\columnwidth}{!}{
\begin{tabular}{lccccccc}
\toprule

 & \thead{Cycle PSNR$\uparrow$}
 & \thead{Multi-View$\uparrow$}
 & \thead{$\mathrm{Abs}(\vect{t})$$\downarrow$}
 & \thead{$\mathrm{Rel}(\vect{t})$$\downarrow$}
 & \thead{$\mathrm{Rel}(\mat{R})$$\downarrow$} \\
 
\midrule
No latents, No pose & 17.75 & 5.975 & 21.83 & 12.43 &  0.518  \\
No Cut3R latents & 17.90 & 6.832 & 19.70 & 11.84 &  0.489 \\
No Cut3R pose & 20.70  & 16.08 & 16.93 & 9.460 & 0.467  \\

\midrule
\textbf{Ours} & {20.74}& {17.11} & {14.39} & {7.798} & {0.411}  \\
\bottomrule
\end{tabular}
}
\end{center}

\caption{\textbf{Effect of Source Pose Conditioning.} Performance gains over baseline are predominantly driven by CUT3R conditioning, with source pose conditioning having a relatively minor impact.}

\label{table:rebuttal-ablation}
\end{table}

\begin{figure*}[t]
\centering
\includegraphics[width=1.0\linewidth]{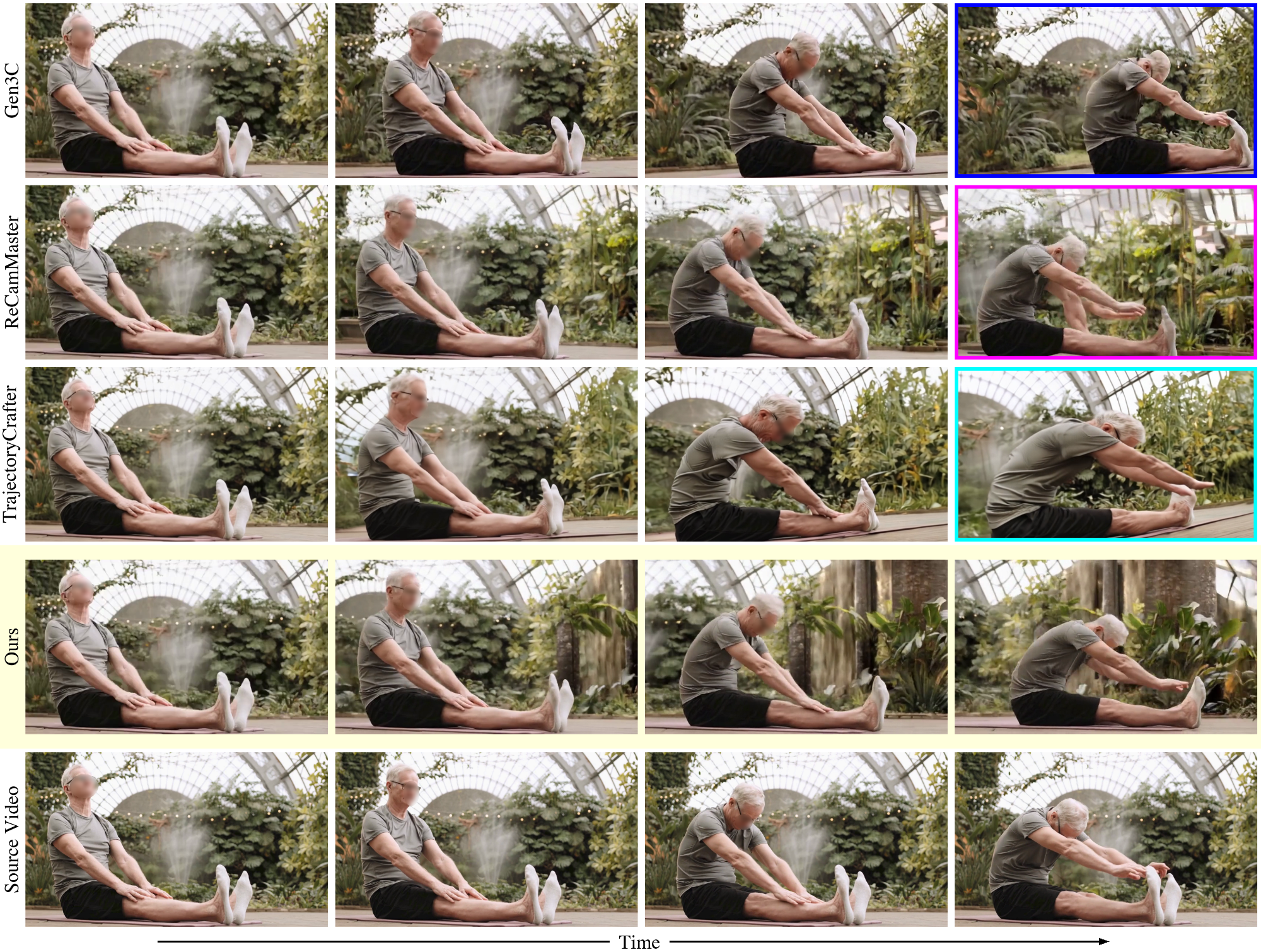}
\caption{\textbf{Qualitative evaluation of novel trajectories across frames.}
Baseline methods conditioned on re-rendered point clouds suffer from unnatural \textcolor{cyan}{\textbf{stretching artifacts and missing details}} (Row 1, Col. 4; Row 3, Col. 4).
The unconditioned baseline \textcolor{magenta}{\textbf{hallucinates a third arm}} (Row 2, Col. 4).
We avoid these pitfalls by using the latent state of a pretrained 4D reconstruction model as a \textit{soft} geometric condition.}%
\label{fig:qualitative-all-grid}
\end{figure*}

\section{Experiments}%
\label{sec:experiments}

\inlineheading{Baselines}
We compare our results with three baseline methods for re-rendering a monocular video along a novel camera trajectory: Gen3C~\cite{ren2025gen3c}, TrajectoryCrafter~\cite{mark2025trajectorycrafter}, and ReCamMaster~\cite{bai2025recammaster}.
The first two use a re-rendered point cloud as a conditioning signal, while the last is unconditioned.
For each baseline, we use the code implementation and pretrained models provided by the authors.
We set camera parameters to ensure all evaluated methods follow similar trajectories.

\inlineheading{Implementation Details}
We use a learning rate of $6\times 10^{-5}$ for the CUT3R adapter and $2\times 10^{-5}$ for the other trainable parameters. Our model has approximately 1.3B parameters, while ReCamMaster~\cite{bai2025recammaster}, TrajectoryCrafter~\cite{yu2024viewcrafter} and Gen3C~\cite{ren2025gen3c} have approximately 1.3B, 5B, and 7B parameters, respectively.

\inlineheading{Training Details}
We train the model on the synthetic MultiCamVideo dataset from ReCamMaster~\cite{bai2025recammaster}, where we randomly select two trajectories per scene as source and target. We train our model on eight H200 GPUs for 15K iterations with a batch size of eight.

\inlineheading{Evaluation Dataset}
To create a diverse evaluation dataset, we obtain a hundred dynamic-scene videos from Pexels (a large stock-footage platform) and fifty static-scene videos from DL3DV~\cite{ling2024dl3dv}.
All videos are resampled to 33 frames and resized to a resolution of 480$\times$832.
For each video, we evaluate all methods under four different novel camera trajectories.
To ensure a fair comparison, each video is accompanied by exactly the same text caption.

\subsection{Results}%
\label{sec:qualitative}

Qualitative results for all baselines and our method are presented in \cref{fig:qualitative-all-methods,fig:qualitative-all-grid}.
Gen3C and TrajectoryCrafter frequently suffer from unnatural warping due to errors in the reconstructed point clouds.
This is clearly illustrated in \cref{fig:qualitative-trajcrafter-errors}, which shows the rendered point cloud condition along with the generated output of TrajectoryCrafter.
ReCamMaster, on the other hand, lacks any geometric conditioning, hallucinates implausible content, and fails to maintain object consistency across occlusions (\cref{fig:qualitative-recam-errors}).
Our method generates views that look more natural and consistent.

\inlineheading{Pose Reconstruction Accuracy}
We evaluate the accuracy of the generated trajectories by running the state-of-the-art dynamic scene bundle adjustment method of Chen\etal~\cite{chen2025back} on the output of each baseline.
We then use the Umeyama algorithm~\cite{umeyama2002least} to align the predicted poses with the ground truth targets and calculate average absolute and relative errors over all frames.
We present the results in \cref{table:quantitative-results-d}.
Our method closely follows the target trajectory, with the lowest translation and rotation errors of all baselines.

\inlineheading{Cycle Consistency}
To measure the consistency of the generated videos, we evaluate the symmetry between generated frames along a cyclical target trajectory.
For static scenes, the generated views should match when the camera revisits the same pose.
Thus, we evaluate all baselines for cycle consistency using 50 random videos from the DL3DV dataset.
Our method outperforms TrajectoryCrafter and ReCamMaster and has consistently lower error than Gen3C (\cref{table:quantitative-results-c}).

\inlineheading{Video Quality}
We evaluate video generation quality using the VBench 1 \& 2~\cite{huang2023vbench, zheng2025vbench2} suite of metrics on dynamic video inputs (\cref{fig:vbench-metrics,table:quantitative-results-c}).
Our method shows all-round high performance without any large failures and achieves the best results on all consistency metrics.

\subsection{Qualitative Results on 4D Reconstruction}
\begin{figure}[h!]
\vspace{-5pt}
  \centering
  \includegraphics[width=\linewidth]{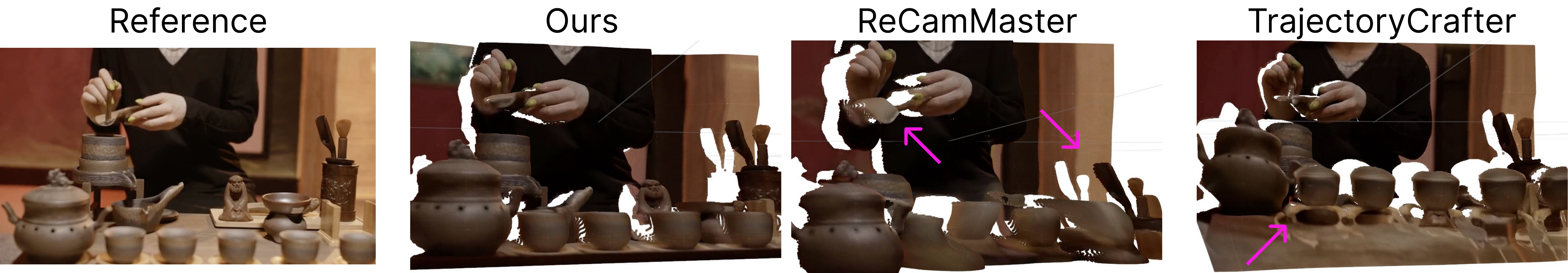}
    \vspace{-12pt}

  \caption{\textbf{Qualitative Results on 4D Reconstruction.} We run BA-Track~\cite{chen2025back} on re-rendered videos and visualize the 4D point clouds from a novel view. Ours has the the least amount of hallucination.}
  \label{fig:rebuttal-4drecon}
\vspace{-2mm}

\end{figure}

 While the generative nature of this task makes it impossible to identify a single ground truth for direct geometric comparison, we run 4D reconstruction~\cite{chen2025back} on the re-rendered videos and compare the results qualitatively in Fig.~\ref{fig:rebuttal-4drecon}. The 4D scene reconstructed from our method exhibits the least amount of hallucination overall.

\vspace{-1mm}


\subsection{Ablation on Source Pose Conditioning}
\label{sec:ablation}

Our model is conditioned on both scene latent tokens and source camera poses. 
In Tab.~\ref{table:rebuttal-ablation}, we show that conditioning on scene latent tokens alone yields a substantial improvement over the baseline, while adding source pose conditioning provides an additional but smaller gain. This indicates that latent token conditioning is the primary contributor, with pose conditioning playing a complementary role.

\paragraph{Limitations}
Our method currently struggles with moving transparent objects (e.g., a glass cup being lifted by a person), likely due to CUT3R's limited ability to estimate reliable geometry for such scenes. Moreover, our CUT3R conditioning mechanism also incurs additional computation cost.

\section{Conclusion}%
\label{sec:conclusion}

We present a method for novel camera trajectory synthesis in dynamic scenes by conditioning a video diffusion model on latents from a large 4D reconstruction model (LRM).
These latents provide geometry-aware guidance in a ``soft'' form, allowing the pretrained diffusion prior to regularize local inconsistencies and avoid the errors and rigidity of rendered point clouds.
Experiments on both static and dynamic scenes demonstrate that our approach achieves stronger geometric consistency and higher visual quality than existing geometry-conditioned and unconditioned baseline methods.

\vspace{54pt}

\section{Acknowledgement}%
\label{sec:acknowledgement}
\vspace{7pt}

We sincerely thank Bo Zhu and Denis Demandolx for their support and valuable discussions that improved this work. We sincerely thank Yixuan Ren for his  insightful technical suggestions, particularly on model design. \newline
We also thank Hanyu Wang and Luxi Zhao for providing useful feedback on our method's qualitative results.

\newpage
{
    \small
    \bibliographystyle{ieeenat_fullname}
    \bibliography{main}
}

\end{document}